\documentclass[final]{cvpr}

\usepackage{times}
\usepackage{epsfig}
\usepackage{graphicx}
\usepackage{amsmath}
\usepackage{amssymb}
\usepackage{booktabs}
\usepackage{subfigure}
\usepackage{wrapfig}

\usepackage{array}
\usepackage{bm}
\usepackage{multirow}
\usepackage[table,xcdraw]{xcolor}
\usepackage{diagbox}
\usepackage{rotfloat}
\newcolumntype{x}{>{\columncolor[HTML]{EFEFEF}[16pt]}r}
\newcolumntype{y}{>{\columncolor[HTML]{EFEFEF}[0pt]}l}
\newcommand{\mainaccuracy}[1]{\multicolumn{1}{x@{}}{#1}}
\newcommand{\plusaccuracy}[1]{\multicolumn{1}{@{}y}{\footnotesize{\color{red}#1}}}
\usepackage{algorithm}
\usepackage{algpseudocode}
\algnewcommand\algorithmicinput{\textbf{Input:}}
\newlength{\maxwidth}
\newcommand{\algalign}[2]
{\makebox[\maxwidth][l]{$#1{}$}${}#2$}

\usepackage{multicol}
\usepackage{setspace}
\usepackage{caption}

\usepackage[pagebackref=true,breaklinks=true,colorlinks,bookmarks=false]{hyperref}

\def\cvprPaperID{2806} 
\def\confYear{CVPR 2021}

\begin{document}

\title{Distilling Causal Effect of Data in Class-Incremental Learning}

\author{
Xinting Hu\textsuperscript{1},\quad Kaihua Tang\textsuperscript{1}, \quad Chunyan Miao\textsuperscript{1},\quad Xian-Sheng Hua\textsuperscript{2}, \quad Hanwang Zhang\textsuperscript{1}\\
{\small \textsuperscript{1}Nanyang Technological University,\quad \textsuperscript{2}Damo Academy, Alibaba Group}\\
{\tt \small xinting001@e.ntu.edu.sg, \quad kaihua001@e.ntu.edu.sg, } \\
{\tt \small ascymiao@ntu.edu.sg, \quad xiansheng.hxs@alibaba-inc.com, \quad hanwangzhang@ntu.edu.sg}\\

}

\maketitle

\begin{abstract}
We propose a causal framework to explain the catastrophic forgetting in Class-Incremental Learning (CIL) and then derive a novel distillation method that is orthogonal to the existing anti-forgetting techniques, such as data replay and feature/label distillation. We first 1) place CIL into the framework, 2) answer why the forgetting happens: the causal effect of the old data is lost in new training, and then 3) explain how the existing techniques mitigate it: they bring the causal effect back. Based on the framework, we find that although the feature/label distillation is storage-efficient, its causal effect is not coherent with the end-to-end feature learning merit, which is however preserved by data replay. To this end, we propose to distill the Colliding Effect between the old and the new data, which is fundamentally equivalent to the causal effect of data replay, but without any cost of replay storage. Thanks to the causal effect analysis, we can further capture the Incremental Momentum Effect of the data stream, removing which can help to retain the old effect overwhelmed by the new data effect, and thus alleviate the forgetting of the old class in testing. Extensive experiments on three CIL benchmarks: CIFAR-100, ImageNet-Sub\&Full, show that the proposed causal effect distillation can improve various state-of-the-art CIL methods by a large margin (0.72\%--9.06\%). \footnote{Codes is available at \url{https://github.com/JoyHuYY1412/DDE_CIL}}
\end{abstract}
\section{Introduction}\label{1}
Any learning systems are expected to be adaptive to the ever-changing environment. In most practical scenarios, the training data are streamed and thus the systems cannot store all the learning history: for animate systems, the neurons are genetically designed to forget old stimuli to save energy~\cite{brainforget2,brainforget1}; for machines, old data and parameters are discarded due to limited storage, computational power, and bandwidth~\cite{lifelong2,lifelong1,Zhang_2021_CVPR}. In particular, we prototype such a practical learning task as Class-Incremental Learning\footnote[1]{There are also other settings like task-incremental~\cite{TIL2,TIL1}.} (CIL)~\cite{e2e,lucir,icarl,bic}, where each incremental data batch consists of new samples and their corresponding new class labels. The key challenge of CIL is that discarding old data is especially catastrophic for deep-learning models, as they are in the data-driven and end-to-end representation learning nature. The network parameters tuned by the old data will be overridden by the SGD optimizer using new data~\cite{SGD1,SGD2}, causing a drastic performance drop --- catastrophic forgetting~\cite{cat_forget1,cat_forget2,lifelong2} --- on the old classes.

\begin{figure}[t]
\setlength{\abovecaptionskip}{0.cm}
\setlength{\belowcaptionskip}{-0.cm}
\begin{center}
  \includegraphics[width=0.82\linewidth]{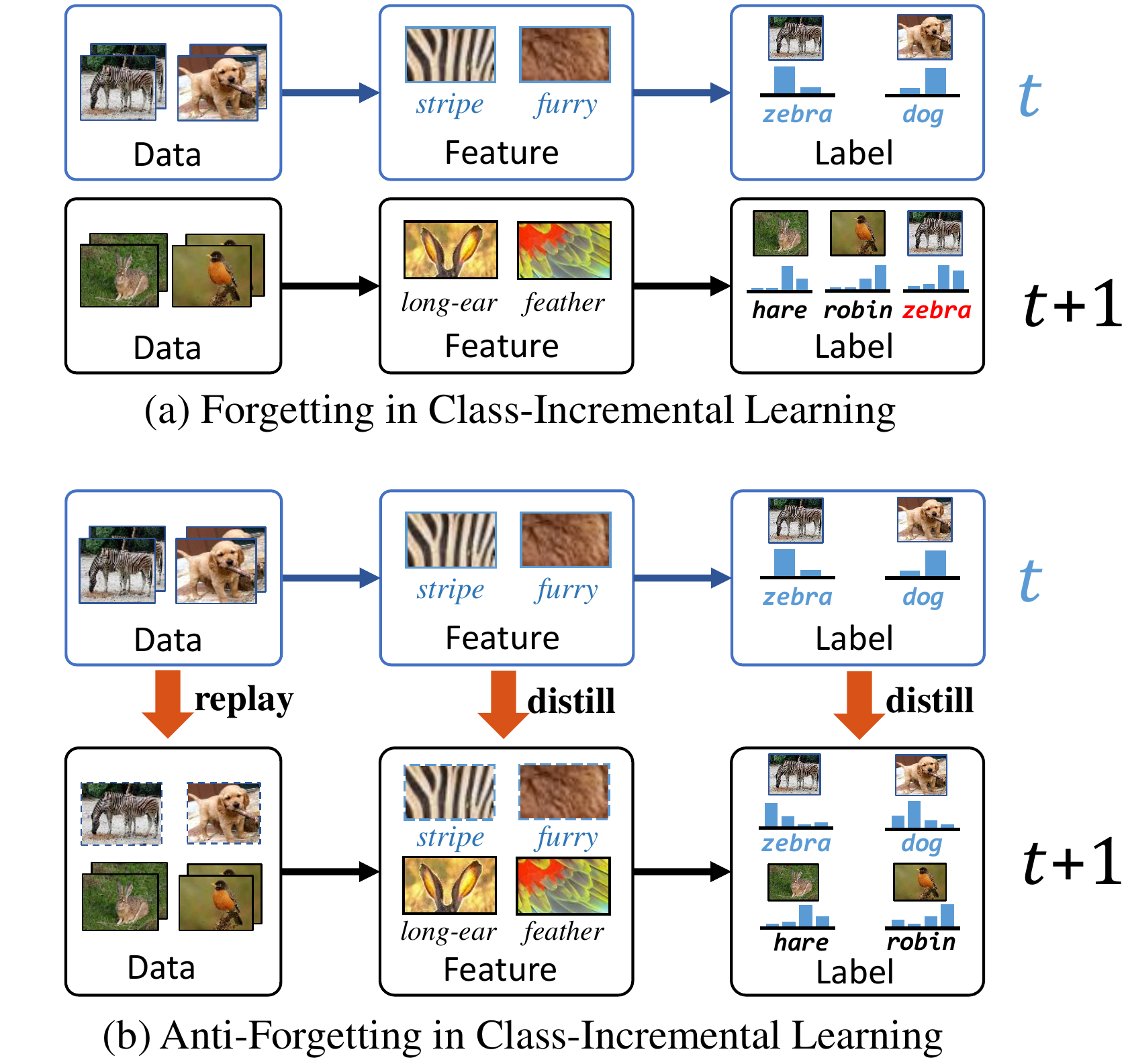}
\end{center}
\vspace{-0.1in}
  \caption{Forgetting and existing anti-forgetting solutions in Class-Incremental Learning at step $t$ and $t+1$. (a): Forgetting happens when the features are overridden by the $t+1$ step training only on the new data. (b): The key to mitigate forgetting is to impose data, feature, and label effects from step $t$ on step $t+1$.}
\vspace{-0.25in}
\label{fig:tf}
\end{figure}

We illustrate how CIL forgets old classes in Figure~\ref{fig:tf} (a). At step $t$, the model learns to classify \textit{``dog''} and \textit{``zebra''}; at the next step $t+1$, it updates with new classes \textit{``hare''} and \textit{``robin''}. If we fine-tune the model only on the new data, old features like ``stripe'' and ``furry'' will be inevitably overridden by new features like ``long-ear'' and ``feather'', which are 
discriminative enough for the new classes. Therefore, the new model will lose the discriminative power --- forget --- about the old classes (\eg, the red ``\textit{zebra}'' at step $t+1$). 

To mitigate the forgetting, as illustrated in Figure~\ref{fig:tf} (b), existing efforts follow the three lines:  1) data replay~\cite{icarl,lucir,mne}, 2) feature distillation~\cite{lucir, pod}, and 3) label distillation~\cite{e2e,bic}. Their common nature is to impose the old training \emph{effect} on the new training (the red arrows in Figure~\ref{fig:tf} (b), which are formally addressed in Section~\ref{sec:3.3}). To see this, for replay, the imposed effect can be viewed as including a small amount of old data (denoted as dashed borders) in the new data; for distillation, the effect is the features/logits extracted from the new data by using the old network,  which is imposed on the new training by using the distillation loss, regulating that the new network behavior should not deviate too much from the old one.


However, the distillation is not always coherent with the new class learning --- it will even play a negative role at chances~\cite{tpcil, tpcil2}. For example, if the new data are out-of-distribution compared to the old one, the features of the new data extracted from the old network will be elusive, and thus the corresponding distillation loss will mislead the new training~\cite{ood1, ood2}. We believe that the underlying reason is that the feature/label distillation merely imposes its effect at the output (prediction) end, 
violating the end-to-end representation learning, which is however preserved by data replay. In fact, even though that the ever-increasing extra storage in data replay contradicts with the requirement of CIL, it is still the most reliable solution for anti-forgetting~\cite{icarl,replay_better1,replay_better2}. Therefore, we arrive at a dilemma: the end-to-end effects of data replay are better than the output-end effects of distillation, but the former requires extra storage while the latter does not. We raise a question: \emph{is there a ``replay-like'' end-to-end effect distillation}?


In this paper, we positively answer this question by framing CIL into a causal inference framework~\cite{pearl2016causal,causal}, which models the causalities among data, feature, and label at any consecutive $t$ and $t+1$ learning steps (Section~\ref{sec:3.1}). Thanks to it, we can explain the reasons 1) why the forgetting happens: the \emph{causal effect} from the old training is lost in the new training (Section~\ref{sec:3.2}), and 2) why data replay and feature/label distillation are anti-forgetting: they win the effect back (Section~\ref{sec:3.3}). Therefore, the above question can be easily reduced to a more concrete one: \emph{besides replay, is there another way to introduce the causal effect of the old data?} Fortunately, the answer is yes, and we propose an effective yet simple method called: Distilling Colliding Effect (Section~\ref{sec:collider}), where the desired end-to-end effect can be distilled without any replay storage. Beyond, we find that the imbalance between the new data causal effect (\eg, hundreds of samples per new class) and the old one (\eg, 5--20 samples per old class) causes severe model bias on the new classes. To this end, we propose an Incremental Momentum Effect Removal method to remove the biased data causal effect that causes forgetting (Section~\ref{sec:momentum}). 

Through extensive experiments on CIFAR-100~\cite{cifar100} and ImageNet~\cite{imagenet}, we observe consistent performance boost by using our causal effect distillation, which is agnostic to methods, datasets, and backbones. For example, we improve the two previously state-of-the-art methods: LUCIR~\cite{lucir} and PODNet~\cite{pod}, by a large margin (0.72\%--9.06\%) on both benchmarks. In particular, we find that the improvement is especially larger when the replay of the old data is fewer. For example, the average performance gain is 16.81\% without replay, 7.05\% with only 5 samples per old class, and 1.31\% with 20 samples per old class on CIFAR-100. The results demonstrate that our distillation indeed preserves the causal effect of data.

\section{Related Work}
\label{sec:related_work}
\noindent\textbf{Class Incremental Learning (CIL)}. Incremental learning aims to continuously learn by accumulating past knowledge~\cite{inc1,inc2,liu2020meta}.  Our work is conducted on CIL benchmarks, which need to learn a unified classifier that can recognize all the old and new classes combined. Existing works on tackling the forgetting challenge in CIL can be broadly divided into two branches: replay-based and distillation-based, considering the data and representation, respectively. \textit{Replay-based} methods include a small percentage of old data in the new data. Some works~\cite{icarl, e2e, lucir, bic, Lee_2019_ICCV} tried to select a representative set of samples from the old data, and others~\cite{mne,gen_data1,gen_data2,gen_data3} used synthesized exemplars to represent the distribution of the old data. \textit{Distillation-based} methods combine the regularization loss with the standard classification loss to update the network. The regularization term is calculated between the old and new networks to preserve previous knowledge when learning new data~\cite{distillation,distillation2}. In practice, they enforce features~\cite{lucir, pod} or predicted label logits~\cite{e2e, bic, mne} of the new model to be close to those of the old model. \textit{Our work} aims to find a ``data-free'' solution to distill the effect of old data without storing them. Our work is orthogonal to previous methods and brings consistent improvement to them.

\noindent\textbf{Causal Inference}~\cite{pearl2016causal, causal3}. It has been recently introduced to various computer vision fields, including feature learning~\cite{causal_classification,wang2020visual}, few-shot classification~\cite{yue2020interventional}, long-tailed recognition~\cite{tang2020longtailed}, semantic segmentation~\cite{causal_semantic} and other high-level vision tasks\cite{tang2020unbiased, qi2020two, yang2020deconfounded, niu2020counterfactual}. The merit of using causal inference in CIL is that it can formulate the causal effect of all the ingredients, point out the forgetting is all about the vanishing old data effect, and thus anti-forgetting is to retrieve it back.




\section{(Anti-) Forgetting in Causal Views}
\label{sec:3}
\subsection{Causal Graphs}
\label{sec:3.1}

\begin{figure}[t]
\setlength{\abovecaptionskip}{0.cm}
\setlength{\belowcaptionskip}{-0.cm}
\begin{center}
  \includegraphics[width=0.90\linewidth]{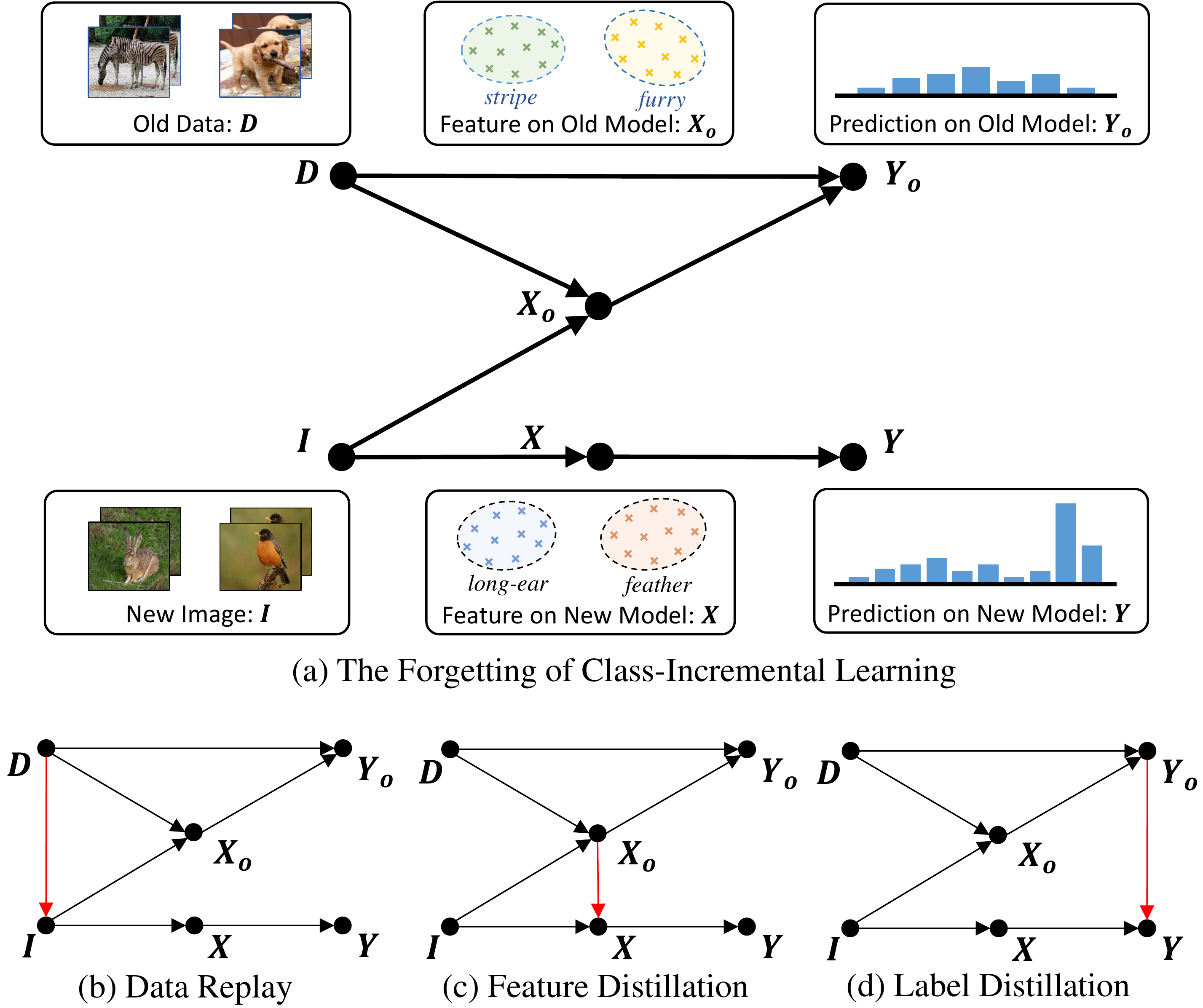}
\end{center}
\vspace{-0.05in}
  \caption{The proposed causal graphs explaining the forgetting and anti-forgetting in CIL. We illustrate the meaning of each node in the CIL framework in (a). The comparison between (a)-(d) shows the key to combating forgetting is the causal effect of old data.}
\vspace{-0.2in}
\label{fig:scm}
\end{figure}

To systematically explain the (anti-)forgetting in CIL in terms of the causal effect of the old knowledge, we frame the data, feature, and label in each incremental learning step into causal graphs~\cite{causal} (Figure~\ref{fig:scm}). The causal graph is a directed acyclic Bayesian graphical model, where the nodes denote the variables, and the directed edges denote the causality between two nodes. 

Specifically, we denote the old data as $D$;  the training samples in the new data as $I$; the extracted features from the current and old model as $X$ and $X_o$, respectively; the predicted labels from the current and old model as $Y$ and $Y_o$, respectively. The links in Figure~\ref{fig:scm} are as follows:

\noindent \bm{$I \rightarrow X \rightarrow Y.$} 
$I \rightarrow X$ denotes that the feature $X$ is extracted from the input image $I$ using the model backbone, $X \rightarrow Y$ indicates that the label $Y$ is predicted by using the feature $X$ with the classifier.

\noindent \bm{$(D,I) \rightarrow X_o\ \&\ (D, X_o) \rightarrow Y_o.$} 
Given the new input image $I$, using the old model trained on old data $D$, we can get the feature representation $X_o$ in the old feature space. Similarly, the predicted logits $Y_o$ in the old class vocabulary can be obtained by using the feature representation $X_o$ and old classifier trained on $D$. These two causal sub-graphs denote the process of interpreting each new image $I$  using the previous model trained on old data $D$. 

\noindent \bm{$D \rightarrow I.$}
 As the data replay strategy stores the old data and mixes them with the new, this link represents the process of borrowing a subset of the old data from $D$, and thus affecting the new input images $I$.

\noindent \bm{$X_o \rightarrow X\ \&\ Y_o \rightarrow Y.$} The old feature representation $X_o$ (old logits $Y_o$) regularizes the new feature $X$ (new logits $Y$) by the feature (logits) distillation loss.

\noindent \bm{$X_o \not\rightarrow X.$} Though the new model is initialized from the old model, the effect of initial parameters adopted from the old model will be exponentially decayed towards $0$ during learning~\cite{overcome}, and thus is neglected in this case.

\subsection{Forgetting} 
\label{sec:3.2}
Based on the graph, we can formulate the causal effect between variables by \emph{causal intervention}, which is denoted as the $do(\cdot)$ operation~\cite{causal2,pearl2016causal}. $do(A=a)$ is to forcibly and externally assign a certain value $a$ to the variable $A$, which can be considered as a ``surgery'' on the graph by removing all the incoming links of $A$, making its value independent of its parents (causes). After the surgery, for any node pair $A$ and $B$, we can estimate the effect of $A$ on $B$ as $P(B\mid \! do(A\!=\!a)) - P(B\mid\!do(A\!=\!0))$, where $do(A\!=\!0)$ denotes the null intervention, \eg, the baseline compared to $A = a$.

To investigate why the forgetting happens in the new training process, we consider the difference between the predictions $Y$ with and without the existence of old data $D$. For each incremental step in CIL, the effect of old data $D$ on $Y$ for each image is formulated as: 
\begin{align}
\textit{Effect}_D & = {P}(Y\!=\!y\mid do(D\!=\!d)) - {P}(Y\!=\!y\mid do(D\!=\!0)) \nonumber\\
&={P}(Y\!=\!y\mid  D\!=\!d) - {P}(Y\!=\!y\mid  D\!=\!0) ,
\label{eq:causal_effect_k}
\end{align}
where  ${P}(Y| do(D))={P}(Y| D)$ since $D$ has no parents. 

As observed in Figure~\ref{fig:scm}~(a), all the causal paths from $D$ to $Y$ is blocked by colliders. Interested readers may jump to the first paragraph of Section~\ref{sec:4.1} to know the collider property. For example, the path $D\rightarrow X_o\leftarrow I\rightarrow X\rightarrow Y$ is blocked by the collider $X_o$. Therefore, we have:
\begin{align}
\textit{Effect}_D & = {P}(Y\!=\!y\mid  D\!=\!d) - {P}(Y\!=\!y\mid  D\!=\!0) \nonumber\\
& =P(Y=y) - P(Y=y) = 0.
\label{eq:zero_effect_D}
\end{align} 
The zero effect indicates that $D$ has no influences on $Y$ in the new training --- forgetting.

\subsection{Anti-Forgetting}
\label{sec:3.3}
Now we calculate the causal effect $\textit{Effect}_D$ when using existing anti-forgetting methods: data replay and feature/label distillation as follows:

\noindent\textbf{Data Replay}. Since the data replay strategy introduces a path from $D$ to $I$ as in Figure~\ref{fig:scm} (b), $D$ and $I$ is no longer independent with each other. By expanding Eq.~\eqref{eq:causal_effect_k}, we have:
\begin{align}
&\textit{Effect}_D \!=\! \sum\nolimits_{I}{P(Y\mid I, D=d)\ P(I\mid D=d)}\!\nonumber\\
                  &-\!\sum\nolimits_{I}{P(Y\mid I, D=0)\ P(I\mid D=0)} \nonumber\\
      &=\sum\nolimits_{I}{P(Y\!\mid\! I)~\left( P(I\!\mid\! D\!=\!d)- P(I\!\mid\! D\!=\!0)\right)}\neq 0,
\label{eq:unfold_probability_replay}
\end{align}
where $P(I\mid D\!=\!d)\!-\!P(I\mid D\!=\!0)\neq 0$ due to the fact that the replay changes the sample distribution of the new data, and $P(Y\mid I,D)\!=\!P(Y\mid I)$ as $I$ is the only mediator from $D$ to $Y$. It is worth noting that as the first term $P(Y\mid I,D)$ is calculated in an end-to-end fashion: from the data $I$ to the label prediction $Y$, data replay takes the advantages of the end-to-end feature representation learning.

\noindent\textbf{Feature \& Label Distillation}. Similarly, as the causal links $X_o \rightarrow X$ and $ Y_o \rightarrow Y$ are introduced into the graph by using feature and logits distillation, the path $D\rightarrow X$ and $D\rightarrow Y$ are unblocked. Take the feature distillation for example, substituting $X$ for $I$ in Eq.~\eqref{eq:unfold_probability_replay}, we have 
\begin{equation}
\resizebox{0.9\hsize}{!}{$
\textit{Effect}_D\!=\!\sum\limits_{X}\!{P(Y\!\mid\!X)~( P(X\!\mid\!D\!=\!d)\!-\! P(X\!\mid\!D\!=\!0))}\neq 0,
\label{eq:unfold_probability_distill}
$}
\end{equation}
where $P(X\mid\!D\!=\!d)\!-\! P(X\mid\!D\!=\!0)\!\neq\!0$ because the old network regularizes the new feature $X$ via the distillation. Note that $P(Y\mid X)$ is not end-to-end as that in data replay.

Thanks to the causal graphs in Figure~\ref{fig:scm}, we explain how the forgetting is mitigated by building unblocked causal paths between $D$ and $(I,X,Y)$, and then introduce the non-zero $\textit{Effect}_D$. Most of the recent works in CIL can be considered as the combinations of these cases in Figure~\ref{fig:scm} (b)-(d). For example, iCaRL~\cite{icarl} is the combination of (b) and (d), where old exemplars are stored and the distillation loss is calculated from soft labels. LUCIR~\cite{lucir}, which adopts the data replay and applies a distillation loss on the features while fixing old class classifiers, can be viewed as an integration of (b), (c), and (d).

\section{Distilling Causal Effect of Data}
\label{sec:4.1}

Thanks to the causal views, we first introduce a novel distillation method that fundamentally equals to the causal effect of data replay, but without any cost of replay storage (Section~\ref{sec:collider} and Algorithm~\ref{alg:4.1}). Then, we remove the overwhelmed new data effect to achieve balanced prediction for the new and old classes (Section~\ref{sec:momentum} and Algorithm~\ref{alg:4.2}).

\begin{algorithm}
\setstretch{1.1}
\caption{\small One CIL step with Colliding Effect Distillation}
\begin{algorithmic}[1] 
    \settowidth{\maxwidth}{$mmmm$}
    \State \algalign{\textbf{Input}}{:}\ $ \mathcal{I},\  \it{\Omega}_o$ \Comment{new training data, old model}
    \State \algalign{\textbf{Output}}{:}\ $ \it{\Omega}$ \Comment{new model}
    \State $      \it{\Omega}\ \ \gets\ \it{\Omega}_o$ \Comment{initialize new model}
    \State $\mathcal{X}_o \gets \it{\Omega}_o(\mathcal{I})$ \Comment{represent new images in old features}
    \Repeat \ for\ $I \in \mathcal{I}$  
     \State ${ \mathcal{N}_K \ \gets\ \Call{K-Nearest-Neighbor}{\it{\Omega_o}(I),\mathcal{X}_o} } $ 
     \State $ P(Y\!\mid\!N_1), \dots, P(Y\!\mid\!N_K) \gets \it{\Omega}(\mathcal{N}_K)$
     \State $W_1,\dots,W_K \gets \Call{WeightAssign}{K}$ \Comment{Eq.~\eqref{eq:weight_assignment}}
     \State $\it{Effect} \gets \sum_{j=1}^{K}{W_j~P(Y\!\mid\!N_j;~\it{\Omega})}$
     \State $\it{\Omega} \gets \mathop{\arg\min} \limits_{\it{\Omega}} (-~\log(\it{Effect}))$
    \Until{converge} 

\end{algorithmic}
\label{alg:4.1}
\end{algorithm}

\begin{algorithm}
\setstretch{1.05}
\caption{\small CIL with Incremental Momentum Effect Removal}
\begin{algorithmic}[1] 
    \settowidth{\maxwidth}{$mmmm$}
    \State \algalign{\textbf{Input}}{:}\ $\mathcal{I}_1, \mathcal{I}_2, \dots,\mathcal{I}_T$  \Comment{training data of $T$ steps}
    \State \algalign{\textbf{Input}}{:}\ $I$  \Comment{a testing image}
    \State \textbf{For}\ $t \in 1,2,\dots, T $ \Comment{each CIL step} 
     \State \quad $\alpha, \beta, h_t \gets \Call{MovingAverage}{\mathcal{I}_t;\ \Omega_{t}}$  \Comment{training}
     \State \quad $h \gets (1-\beta)\ h_{t-1}+\beta\ h_t$ \Comment {head direction}
    \State \quad $X \gets \Call{FeatureExtractor}{I}$ \Comment{inference}
    \State \quad $Y \ \gets \Call{Classifier}{X-{x^h}}$ \Comment{Eq.~\eqref{eq:momentum}}
\end{algorithmic}
\label{alg:4.2}
\end{algorithm}

\vspace{-0.2in}
\subsection{Distilling Colliding Effect}
\label{sec:collider}

As we discussed in Eq.~\eqref{eq:unfold_probability_replay},  to introduce the end-to-end data effect, the path between $D$ and $I$ is linked by replaying old data. However, the increasing storage contradicts the original intention of CIL~\cite{icarl}. Is there another way around to build the path between $D$ and $I$ without data replay?
%



\begin{figure}
\subfigure[ ]{
  \begin{minipage}[t]{0.44\linewidth}
    \centering
    \includegraphics[scale=0.29]{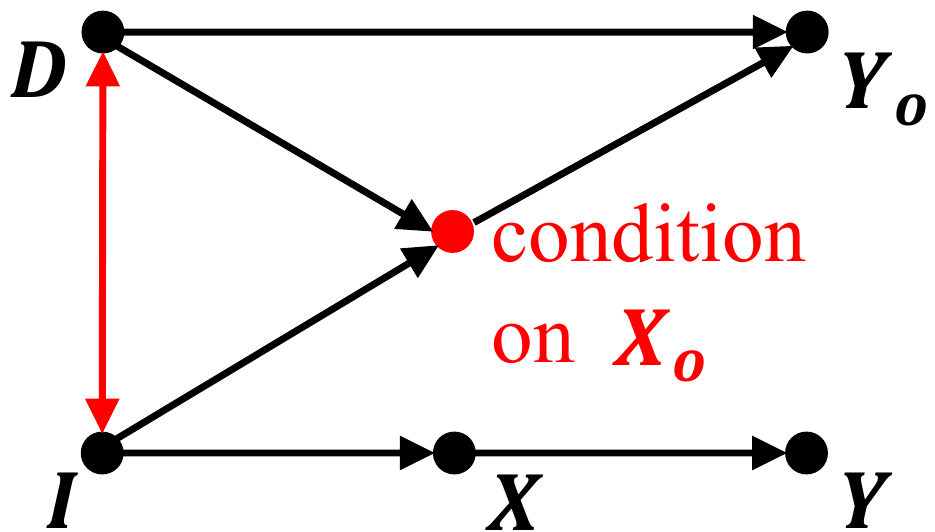}
  \end{minipage}} 
  \subfigure[ ]{
  \begin{minipage}[t]{0.5\linewidth}
    \centering
    \includegraphics[scale=0.38]{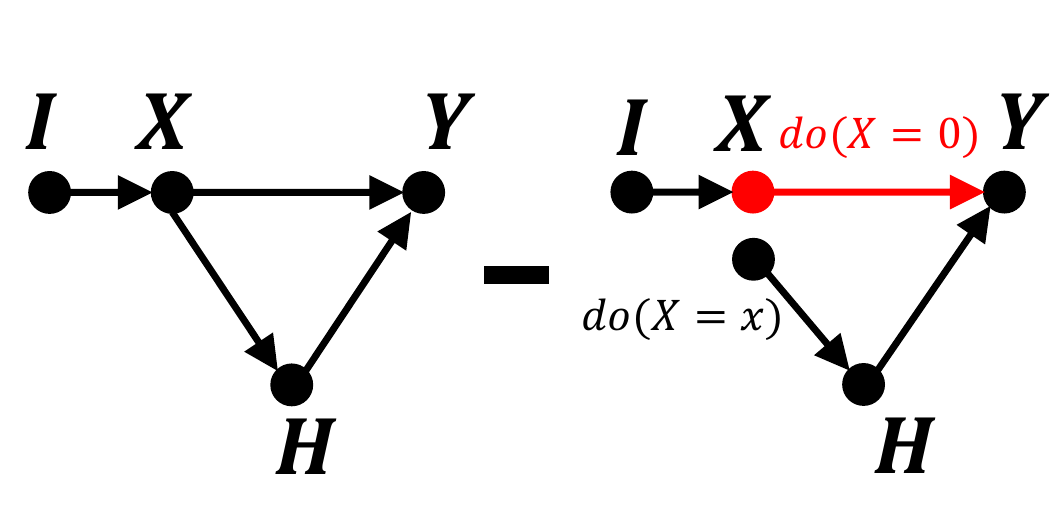}
  \end{minipage}}
  \vspace{-0.1in}
  \caption{(a): The colliding effect in CIL. (b): The causal graph of removing incremental momentum effect.} 
  \label{fig:tde}
  \vspace{-0.2in}
\end{figure}

%
Reviewing the causal graph in Figure~\ref{fig:tf} (a), we observe the silver lining behind the collider $X_o$: If we control the collider $X_o$, \ie, condition on $X_o$, nodes $D$ and $I$ will be correlated with each other, as shown in Figure~\ref{fig:tde} (a). We call this \emph{Colliding Effect}~\cite{pearl2016causal}, which can be intuitively understood through a real-life example: a causal graph \textit{Intelligence} $\rightarrow$ \textit{Grades} $\leftarrow$ \textit{Effort} representing a student’s grades are determined by both the intelligence and hard work. Usually, a student's intelligence is independent of whether she works hard or not. However, considering a student who got good grades and knowing that she is not smart, it tells us immediately that she is most likely hard-working, and vice versa. Thus, two independent variables become dependent upon knowing the value of their joint outcome: the collider variable (\textit{Grades}). 

Based on Eq.~\eqref{eq:causal_effect_k}, after conditioning on $X_o$, the effect of old data can be written as:
{\small
\begin{align}
\textit{Effect}_D\!=\!&\sum_{I}{P(Y\!\mid\! I, X_o)~\underline{(P(I\!\mid\! X_o,\!D\!=\!d) - P(I\!\mid\!X_o,\!D\!=\!0))}}\nonumber\\
\!=\!&\sum_{I}{P(Y\mid I)\ W(I,X_o,D)},
\label{eq:collider_effect}
\end{align}
}%
where $P(Y\!\mid\! I,X_o)=P(Y\!\mid\! I)$ since $I$ is the only mediator from $X_o$ to $Y$, and the underlined term can be denoted as a weight term $W(I,X_o,D)$. By calculating the classification  $P(Y\!\mid\! I\!=\!i)$ and the weight $W(I\!=\!i,X_o,D)$, we can distill the colliding effect into the new learning step without old data replay. 

Intuitively, the value of $W(I=i,X_o,D)$ (written as $W$ below for simplicity) implies the likelihood of one image sample $I\!=\!i$ given $X_o\!=\!x_o$ extracted by model trained on $D\!=\!d$, that is, the larger value of $W$ shows that the more likely the image is like $i$. 
Given all the new images represented by using the old network ($D\!=\!d$), suppose an image $i^*$ whose old feature is $x_o$. According to the meaning of $W$, the value is larger for an image that is more similar to $i^*$. Since the similarity of images can be measured by feature distance, we can infer that the image whose old feature $X_o$ is closer to $x_o$ should have a larger $W$. For the new data containing $n$ image, we can sort all the images according to the distance to $x_o$ ascendingly. The sorted $n$ images are indexed as $\{N_1, N_2,\dots,N_n\}$. 

In summary, the rule of assigning $W$ among these $n$ images can be summarized as: $W_{N_1} \geq W_{N_2} \geq \dots \geq W_{N_n}.$
For implementation, we select a truncated top-$K$ fragment and re-write Eq.~\eqref{eq:collider_effect} as:
\begin{equation}
\textit{Effect}_D\!=\! \sum_{i \in \{N_1, N_2,\dots,N_K\}}{P(Y\mid I=i)~W_{i}} ,
\end{equation}
where $\{W_{N_1},\dots, W_{N_K}\}$ subject to
\begin{equation}
\small
\label{eq:weight_assignment}
\left\{
\begin{aligned}
&W_{N_1} \geq W_{N_2} \geq \dots \geq W_{N_K} \\
&W_{N_1} +  W_{N_2} + \dots + W_{N_K} = 1 \ .
\end{aligned}
\right.
\end{equation}
It means that the classification probabilities are first calculated on each of the $K$ images with the current model and then weighted averaged to obtain the final prediction. For more experiments on the choice of $K$ and weight assignment, see Section~\ref{sec:5.2}.

Compared with data replay that builds $D \rightarrow I$, the merit of the above colliding effect distillation is that it can work with very limited or even without old data, which makes it not just much more efficient, but also applicable to large-vocabulary datasets, where each incremental step can only afford replaying very few samples for one class.

\subsection{Incremental Momentum Effect Removal}
\label{sec:momentum}

\begin{figure}[t]
\centering\includegraphics[width=0.75\linewidth]{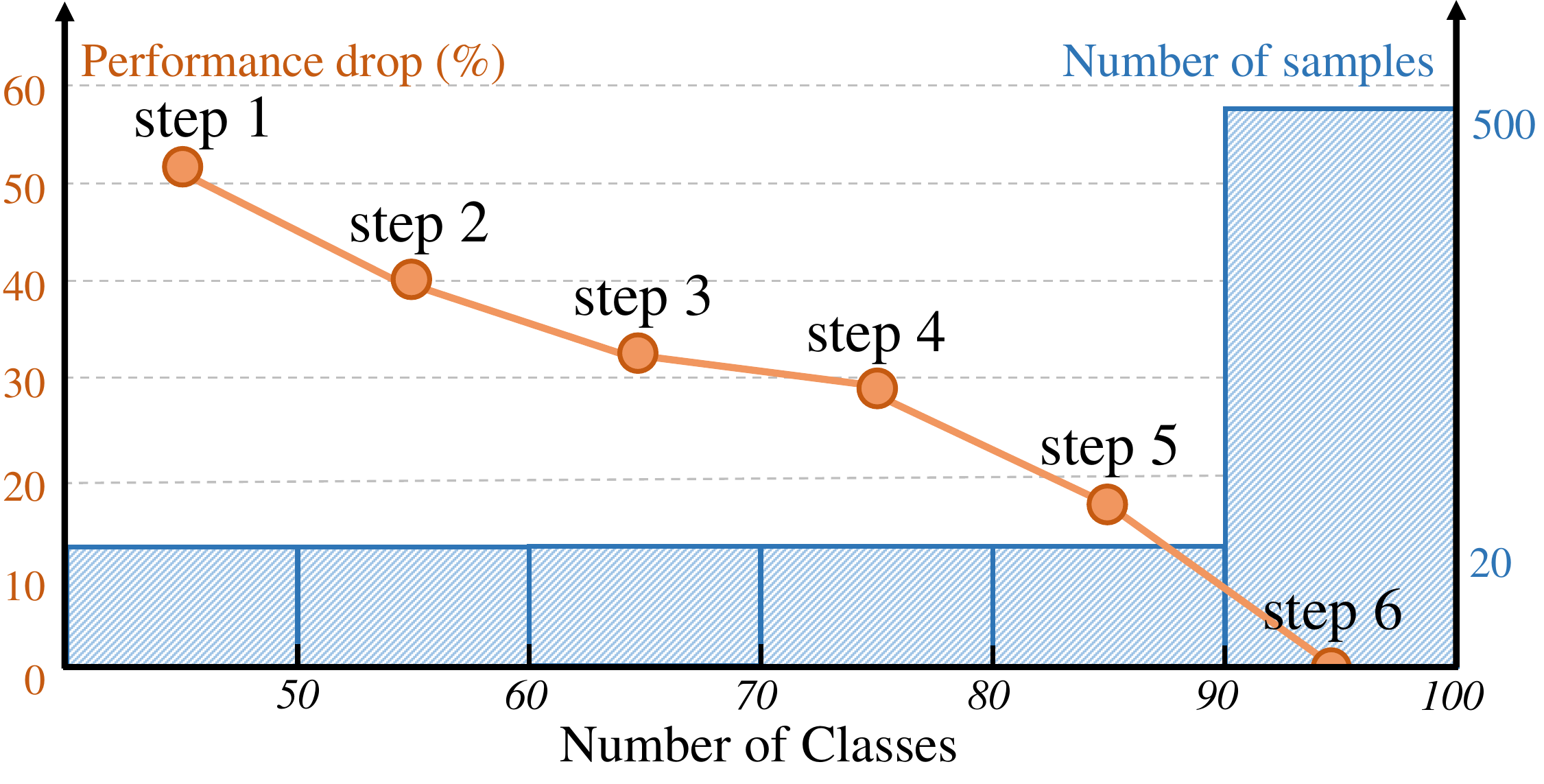}
\vspace{-0.1in}
\caption{The data distribution of CIFAR-100~\cite{cifar100} and the performance drop of learned classes at the final CIL step under the 5-step-20-replay setting.}
\vspace{-0.22in}
\label{fig:long-tail}
\end{figure}

Since the old data barely appear in the subsequent incremental steps, although we ensure the effect between $D$ and $Y$ to be \textit{non-zero}, the moving averaged momentum $\mu$ in SGD optimizer~\cite{sutskever2013importance} inevitably makes the prediction biased towards the new data, causing the CIL system suffers from the severe data imbalance. As shown in Figure~\ref{fig:long-tail}, by visualizing the number of instances for classes in the CIL steps, we observe a clear long-tailed distribution, which is a challenging task.

To mitigate the long-tail bias, we follow~\cite{tang2020longtailed} to pursue the direct causal effect of the feature $X$ during inference, which removes a counterfactual biased head projection from the final logits. The causal graph of this approach is in Figure~\ref{fig:tde} (b), where the nodes $H$ is the head direction caused by the SGD momentum. This causal graph is built on top of Figure~\ref{fig:scm} while only focuses on the current learning process. The subtracted term can be considered as the pure bias effect with a null input. However, the CIL cannot be simply treated as a classic long-tail classification problem because the data distributions across CIL steps are dynamic, \ie, the current rare classes are the frequent classes in a previous step. Our experimental results (Figure~\ref{fig:long-tail}) also reveal that: even the old classes have equally rare frequencies in an incremental step, the performance drop of a class increases with the time interval between the current step and the step when they are initially learned, which is not observed in the long-tail classification. It means that the momentum effect relies not only on the final momentum but also those in previous steps. Therefore, we propose an incremental momentum effect removing strategy. 

For the incremental step $t$ with total $M$ SGD iterations, by moving averaging the feature $x_m$ in each iteration $m$ using $x_m = \mu \cdot x_{m-1} + x_m$ with the SGD momentum $\mu$, we can obtain a head direction $h_t$ as the unit vector $x_M/ \left\| x_M \right \| $ of the final $x_M$. As the moving average of all the samples trained in step $t$, $h_t$ shows the momentum effect of new model. Unlike~\cite{tang2020longtailed} where the $h_t$ is directly adopted, in order to involve the momentum effect of old model, we design a dynamic head direction ${h} = (1-\beta)~h_{t-1} + \beta~h_t$, combining both the head direction of step $t$ and the last step $t-1$. In inference, as illustrated in Algorithm~\ref{alg:4.2}, for each image $i$, we can get the de-biased prediction logits $[Y\!\mid\!I=i]$ by removing the proposed incremental momentum effect as:
\begin{align}
[Y\!\mid\!I=i] & = [Y\!\mid\!do(X\!=\!x)] - \alpha \cdot [Y\!\mid\!do(X\!=\!0, H\!=\!{h})] \nonumber \\
               & = [Y\!\mid\!X\!=\!x] - \alpha \cdot [Y\!\mid\!X\!=\!x^h] ,
\label{eq:momentum}
\end{align} 
where ${x}^h$ means the projection of feature $x$ on the dynamic head direction $h$, and both trade-off parameters $\alpha$ and $\beta$ are learned automatically in the back-propagation.

\section{Experiment} \label{sec:expr}

\begin{table*}[ht]
\setlength{\tabcolsep}{2.8mm}{
  \centering
  \begin{tabular}{ll*{2}{r@{}l}c*{2}{r@{}l}c*{2}{r@{}l}}
  \toprule
  \multicolumn{2}{c}{} & \multicolumn{4}{c}{\textbf{CIFAR-100}} & \textbf{} & \multicolumn{4}{c}{\textbf{ImageNet-Sub}} & \textbf{} & \multicolumn{4}{c}{\textbf{ImageNet-Full}} \\ \cline{3-6} \cline{8-11} \cline{13-16} 
  \multicolumn{2}{l}{\multirow{-2}{*}{\quad \quad \textbf{Methods}}} & \multicolumn{2}{l}{\textit{\quad T=5}} & \multicolumn{2}{l}{\quad \textit{10}} & \textit{} & \multicolumn{2}{l}{\quad \textit{5}} & \multicolumn{2}{l}{\quad \textit{10}} & \textit{} & \multicolumn{2}{l}{\quad \textit{5}} & \multicolumn{2}{l}{\quad \textit{10}} \\ \hline\hline
   & LUCIR$^\dagger$   & 50.76 &  & 53.60 &  &  & 66.44 &  & 60.04 &  &  & 62.61 &  & 58.01 & \\
   & \cellcolor[HTML]{EFEFEF}\quad + DDE (ours) & \mainaccuracy{59.82} & \plusaccuracy{+9.06} &\mainaccuracy{\textbf{60.53}} & \plusaccuracy{+6.93} &\mainaccuracy{} &\mainaccuracy{70.86} & \plusaccuracy{+4.42} &\mainaccuracy{68.30} & \plusaccuracy{+8.26} &\mainaccuracy{} &\mainaccuracy{\textbf{66.18}} & \plusaccuracy{+3.57} &\mainaccuracy{\textbf{62.89}} & \plusaccuracy{+4.88}\\ \cline{2-16} 
   & PODNet$^\dagger$  & 53.38 &  & 55.97 &  &  & 71.43 &  & 64.90 &  &  & 61.01 &  & 55.36 & \\
  \multirow{-4}{*}{\rotatebox{90}{\textit{(1) R=5}}} & \cellcolor[HTML]{EFEFEF}\quad + DDE (ours) & \mainaccuracy{\textbf{61.47}} & \plusaccuracy{+8.09} &\mainaccuracy{60.08} & \plusaccuracy{+4.11} &\mainaccuracy{} &\mainaccuracy{\textbf{74.59}} & \plusaccuracy{+3.16} &\mainaccuracy{\textbf{69.26}} & \plusaccuracy{+4.36} &\mainaccuracy{} &\mainaccuracy{63.15} & \plusaccuracy{+2.14} &\mainaccuracy{58.34} & \plusaccuracy{+2.98} \\ \hline \hline
   & LUCIR$^\dagger$   & 61.68 &  & 58.30 &  &  & 68.13 &  & 64.04 &  &  & 65.21 &  & 61.60 & \\
   & \cellcolor[HTML]{EFEFEF}\quad + DDE (ours) & \mainaccuracy{\textbf{64.41}} & \plusaccuracy{+2.73} &\mainaccuracy{\textbf{62.00}} & \plusaccuracy{+3.70} &\mainaccuracy{} &\mainaccuracy{71.20} & \plusaccuracy{+3.07} &\mainaccuracy{69.05} & \plusaccuracy{+5.01} &\mainaccuracy{} &\mainaccuracy{\textbf{67.04}} & \plusaccuracy{+1.83} &\mainaccuracy{\textbf{64.98}} & \plusaccuracy{+3.38} \\ \cline{2-16} 
   & PODNet$^\dagger$  & 61.40 &  & 58.92 &  &  & 74.50 &  & 70.40 &  &  & 62.88 &  & 59.56 & \\
  \multirow{-4}{*}{\rotatebox{90}{\textit{(2) R=10}}} & \cellcolor[HTML]{EFEFEF}\quad + DDE (ours) & \mainaccuracy{63.40} & \plusaccuracy{+2.00} &\mainaccuracy{60.52} & \plusaccuracy{+1.60} &\mainaccuracy{} &\mainaccuracy{\textbf{75.76}} & \plusaccuracy{+1.26} &\mainaccuracy{\textbf{73.00}} & \plusaccuracy{+2.60} &\mainaccuracy{} &\mainaccuracy{64.41} & \plusaccuracy{+1.53} &\mainaccuracy{62.09} & \plusaccuracy{+2.53} \\ \hline \hline
   & LUCIR$^\dagger$   & 63.57 &  & 60.95 &  &  & 70.71 &  & 67.60 &  &  & 66.84 &  & 64.17 & \\
   & \cellcolor[HTML]{EFEFEF}\quad + DDE (ours) & \mainaccuracy{65.27} & \plusaccuracy{+1.70} &\mainaccuracy{62.36} & \plusaccuracy{+1.41} &\mainaccuracy{} &\mainaccuracy{72.34} & \plusaccuracy{+1.63} &\mainaccuracy{70.20} & \plusaccuracy{+2.60} &\mainaccuracy{} &\mainaccuracy{\textbf{67.51}} & \plusaccuracy{+0.67} &\mainaccuracy{\textbf{65.77}} & \plusaccuracy{+1.60} \\ \cline{2-16} 
   & PODNet$^\dagger$  & 64.70 &  & 62.72 &  &  & 75.58 &  & 73.48 &  &  & 65.59 &  & 63.27 & \\
   & \cellcolor[HTML]{EFEFEF}\quad + DDE (ours) & \mainaccuracy{\textbf{65.42}} & \plusaccuracy{+0.72} &\mainaccuracy{\textbf{64.12}} & \plusaccuracy{+1.40} &\mainaccuracy{} &\mainaccuracy{\textbf{76.71}} & \plusaccuracy{+1.13} &\mainaccuracy{\textbf{75.41}} & \plusaccuracy{+1.93} &\mainaccuracy{} &\mainaccuracy{66.42} & \plusaccuracy{+0.83} &\mainaccuracy{64.71} & \plusaccuracy{+1.44} \\ \cline{2-16} 
   & iCaRL$^\dagger$~\cite{icarl}  & 57.17 &  & 52.27 &  &  & 65.04 &  & 59.53 &  &  & 51.36 &  & 46.72 & \\
   & BiC~\cite{bic}       & 59.36 &  & 54.20 &  &  & 70.07 &  & 64.96 &  &  & 62.65 &  & 58.72 & \\
   & LUCIR~\cite{lucir}   & 63.17 &  & 60.14 &  &  & 70.84 &  & 68.32 &  &  & 64.45 &  & 61.57 & \\
   & Mnemonics~\cite{mne} & 63.34 &  & 62.28 &  &  & 72.58 &  & 71.37 &  &  & 64.54 &  & 63.01 & \\
   & PODNet~\cite{pod}    & 64.83 &  & 63.19 &  &  & 75.54 &  & 74.33 &  &  & 66.95 &  & 64.13 & \\
  \multirow{-10}{*}{\rotatebox{90}{\textit{(3) R=20}}} & TPCIL~\cite{tpcil} & 65.34 &  &  63.58 &  &  & 76.27 &  & 74.81 &  &  & 64.89 &  & 62.88 &  \\
  \bottomrule
  \end{tabular}
  \vspace{-0.1in}
  \caption{Comparisons of average incremental accuracies (\%) on CIFAR100, ImageNet-Sub, and ImageNet-Full under 5/10-step-5/10/20-replay settings with our distillation of data effect (DDE) and state-of-the-art. Models with a dagger $^\dagger$ are produced using their officially released code to assure the fair comparison. See Table~\ref{tab:noreplay} for more results without data replay.}
  \label{tab:sota}
  }
  \vspace{-0.15in}
\end{table*}

\subsection{Settings}
\noindent \textbf{Datasets.}
We evaluated the proposed approach on two popular datasets for class-incremental learning, \ie, CIFAR-100~\cite{cifar100} and ImageNet (ILSVRC 2012)~\cite{imagenet}. Following previous works~\cite{icarl,lucir}, for a given dataset, the classes were first arranged in a fixed random order and then cut into incremental splits that come sequentially. Specifically, ImageNet is used in two CIL settings: ImageNet-Sub~\cite{icarl, lucir, mne} contains the first 100 classes of the arranged 1,000 classes and  ImageNet-Full contains the whole 1,000 classes.

\noindent \textbf{Protocols.}
We followed~\cite{lucir,pod} to split the dataset into incremental steps and manage the memory budget. 1) For splitting each dataset, half of all the classes are used to train a model as the starting point, and the rest classes were divided equally to come in $T$ incremental steps. 
2) For the old data replay after each incremental step, a constant number $R$ of images for each class was stored.
In conclusion, the setting for CIL can be denoted as $T$-step-$R$-replay.

\noindent \textbf{Evaluation Metrics.}
All models were evaluated for each step (including the first step). It can be shown as a trend of $T+1$ classification accuracy curve or be average for all the steps, denoted as \textit{Average Incremental Accuracy}. We also reported the \textit{forgetting} curve following Arslan \textit{et al.}~\cite{forget_rate}. The forgetting at step $t~(t>1)$ is calculated as $F_t = \frac{1}{t-1}\sum_{j=1}^{t-1}f_j^t$, where $f_j^t$ denotes the performance drop of classes first learned in step $j$ after the model has been incrementally trained up to step $t > j$ as:
\begin{equation}
f_j^t=\max \limits_{l\in\{1,\dots,t-1\}} a_{l,j} - a_{t,j}, \forall j<t
\end{equation}
where $a_{i,j}$ is the accuracy of the classes first learned in steps $i$ after the training step $j$.
The lower $f_t$ implies the less forgetting on previous steps, and the \textit{Average Incremental Forgetting} is the mean of $f_t$ for $T$ incremental steps.

\begin{figure*}[t]
\centering\includegraphics[width=0.94\linewidth]{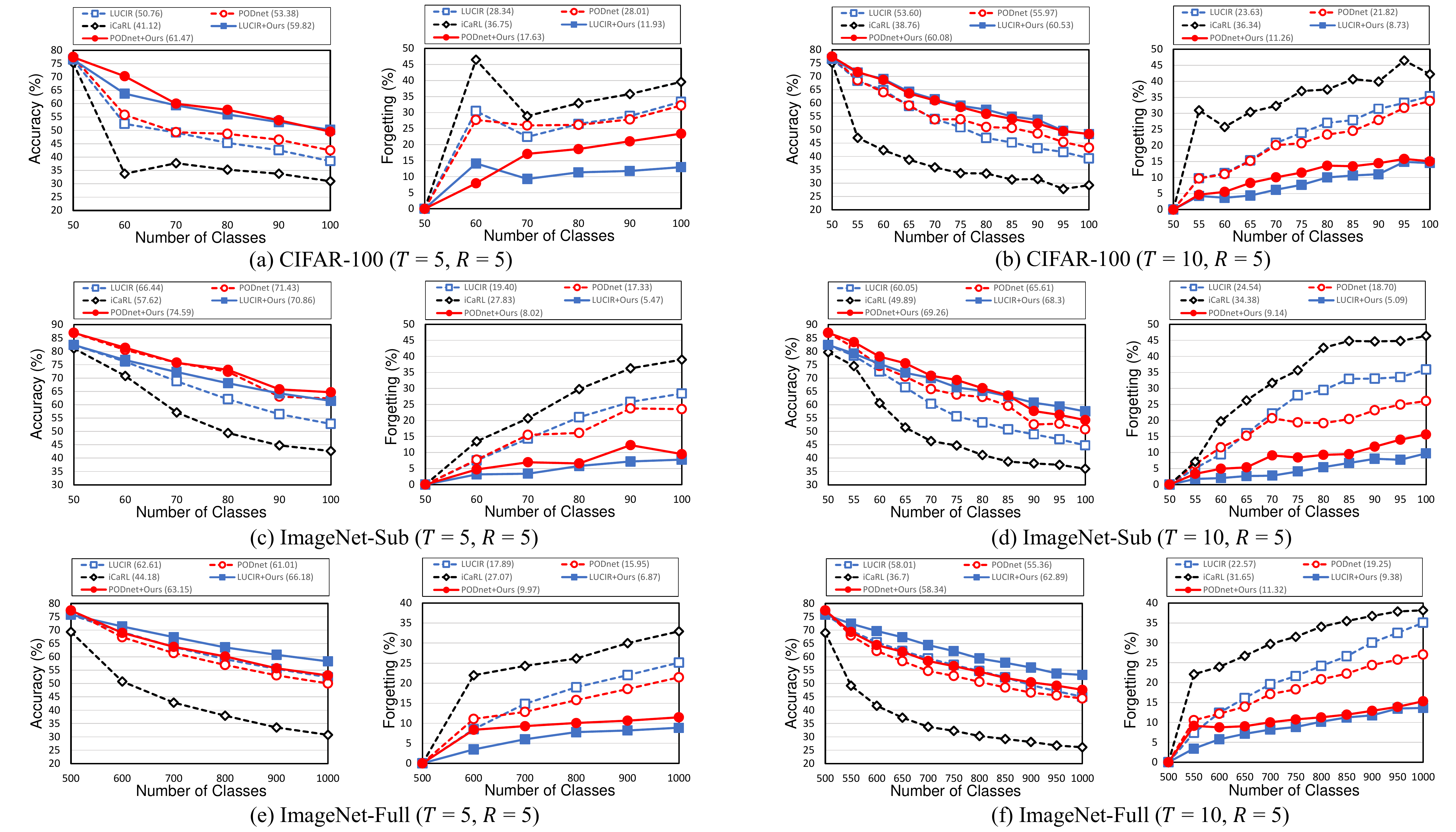}
\vspace{-0.08in}
\caption{Comparisons of the step-wise accuracies and forgettings on CIFAR-100 (100 classes), ImageNet-Sub (100 classes) and ImageNet-Full (1000 classes) when $R=5$.}
\vspace{-0.15in}
\label{fig:curve_res}
\end{figure*}

\noindent \textbf{Implementation Details.}
Following~\cite{lucir, pod}, for the network backbone, we adopted a 32-layer ResNet for CIFAR100, an 18-layer ResNet for ImageNet, and an SGD optimizer with momentum $\mu = 0.9$; for the classifier, a scaled cosine normalization classifier was adopted. Specifically, Douillard \etal~\cite{pod} adopted an ensemble classifier, which is preserved in our experiments on PODNet~\cite{pod}. All models were implemented with PyTorch and run on TITAN-X GPU. 
Since our methods can be implemented as a plug-in module, the common hyper-parameters, \eg, learning rates, batch-sizes, training epochs, herding strategy, and the weights of the losses, were the same as their original settings~\cite{lucir,pod}, which will be given in Appendix. For the proposed Distillation of Colliding Effect (DCE), we used cosine distance to measure the similarity between images. For the number of neighborhood images except for the original image, we adopted 10 for CIFAR100 and ImageNet-Sub, and 1 for ImageNet-Full considering the calculation time. For the incremental Momentum Effect Removal (MER), $\alpha$ and $\beta$ were trained in the finetuning stage using a re-sampled subset containing balanced old and new data. When there is no replay data, we abandoned the finetuning stage and set $\alpha$ and $\beta$ to 0.5 and 0.8 empirically.

\subsection{Results and Analyses} \label{sec:5.2}
\noindent \textbf{Comparisons with State-of-The-Art.}
To demonstrate the effectiveness of our method, we deployed it on LUCIR~\cite{lucir} and PODNet~\cite{pod} using their officially released codes, and compared them with other strong baselines and state-of-the-art. Table~\ref{tab:sota} (3) lists the overall comparisons of CIL performances on the classic 5/10-step-20-replay settings. On CIFAR-100, we can observe that our Distillation of Data Effect (DDE) achieves state-of-the-art 65.42\% and 64.12\% average accuracy, improving the original LUCIR and PODNet up to 1.7\% and surpassing the previous best model~\cite{tpcil} with additional data augmentation. On ImageNet-Sub, we boost LUCIR and PODNet by 1.79\% on average and achieve the new state-of-the-art. For the most challenging ImageNet-Full, our model based on LUCIR achieves 67.51\% and 65.77\%, which surpasses the previous best model by 1.1\% on average. 

\begin{figure}[h]
\centering\includegraphics[width=0.93\linewidth]{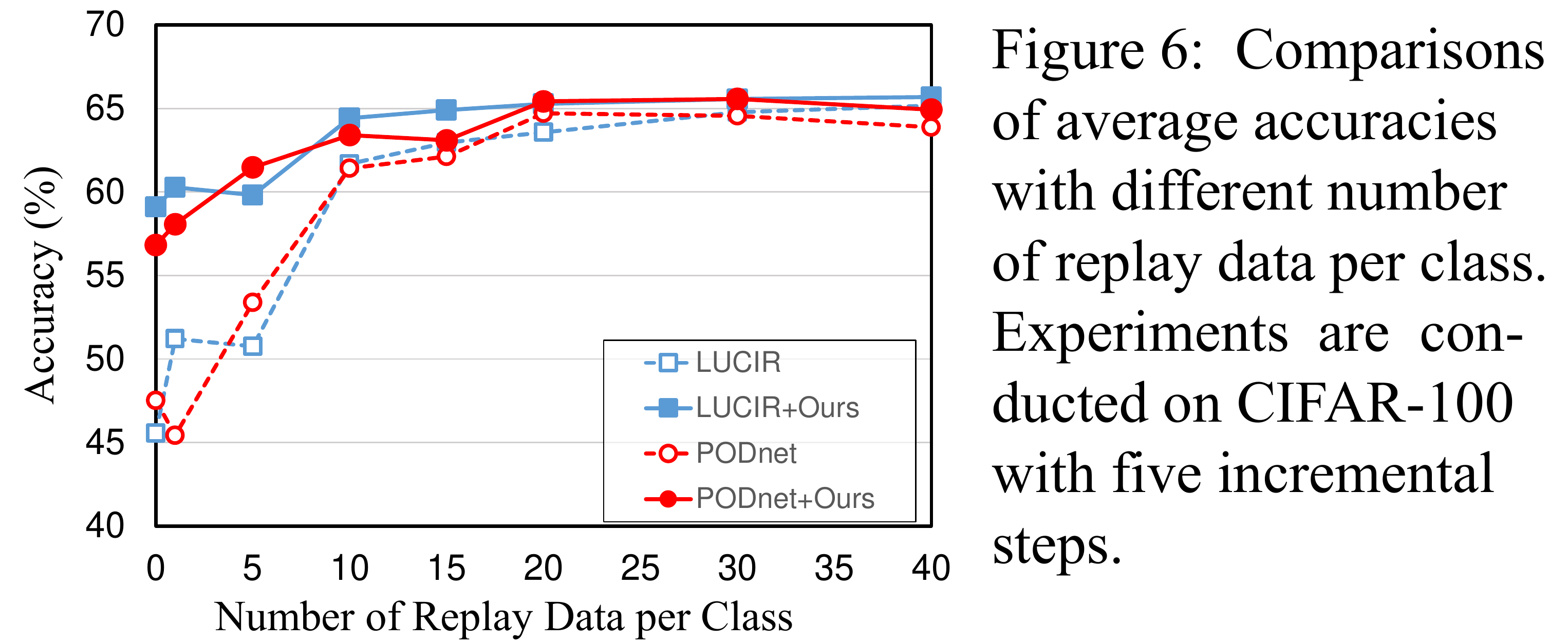}
\captionlistentry{}
\label{fig:vs_data_replay}
\vspace{-0.1in}
\end{figure}

\noindent \textbf{Data Effect Distill vs. Data Replay.} Aiming to distill the data effect, our method can serve as a more storage-efficient replacement for the data-replay strategy. Figure~\ref{fig:vs_data_replay} and Table~\ref{tab:noreplay} illustrate the effect of the data replay compared with our method. It is worth noting that when there is no data replay, \ie, $R\!=\!0$, our method achieves 59.11\% average accuracy, surpassing the result of replaying five samples per class by 13.54\%. To achieve the same result, traditional data-replay methods need to store around 1,000 old images in total, while our method needs nothing. The experiment empirically answers the question we asked initially: our proposed method does distill the end-to-end data effects without replay. Moreover, as the number of old data increases, data-replay based methods improve, while our method still consistently boosts the performance. The results demonstrated the data effect we distilled is not entirely overlapped with the data effect introduced by the old data.

\noindent \textbf{Different Replay Numbers.} To demonstrate the effectiveness and robustness of our data effect, we implemented the model on more challenging occasions where rarer old data are replayed ($R=10,5$). As shown in Table~\ref{tab:sota} (1) and Table~\ref{tab:sota} (2), original methods suffer a lot from the forgetting due to the missing of old data, leading to the rapid decline in the accuracies, \eg, the accuracy drops 12.81\% for CIFAR-100 and 4.23\% for ImageNet as $R$ decreases from 20 to 5. TPCIL also reported a 12.45\% accuracy drop in their ablation study on CIFAR-100~\cite{tpcil}. In contrast, our method shows its effectiveness agnostic to the datasets and the number of replay data. Specifically, our method achieves more improvement when the data constrain is stricter. For example, it obtains up to 9.06\% and 4.88\% improvement on CIFAR-100 and ImageNet when replaying only 5 data per class. To see the step-wise results, we draw Figure~\ref{fig:curve_res} to show the accuracy and forgetting for each learning step. Starting from the same point, our method (in solid) suffers from less forgetting and performs better at each incremental step.

\noindent \textbf{Effectiveness of Each Causal Effect.} Table~\ref{tab:each_component} shows the results of using Distillation of Colliding Effect (DCE) and incremental Momentum Effect Removal (MER) alone and together. One can see that the boost of using each component alone is lower than using them combined. Therefore, they do not conflict with each other and demonstrate that both the causal effects play an important role in CIL.

\noindent \textbf{Different Weight Assignments.} Table~\ref{tab:weight} shows the results of using Distillation of Colliding Effect (DCE) with different weight assignment strategies. \textit{Top-n} means for each image $i$, we select $n$ nearest neighbor images as discussed in Section~\ref{sec:collider} and calculate the average prediction of the images $i$ and the mean of those $n$ images. \textit{Rand} and \textit{Bottom} use randomly selected images or the most dissimilar (farthest) images. \textit{Variant1} and \textit{Variant2} select the same images as \textit{Top 10}, but modify the weight value for those $n+1$ images (See Appendix for detail). \textit{Variant1} lower the weight of the image itself and split more weight to the neighbors, while \textit{Variant2} split the weight in proportion to the softmax of the cosine similarity between images. By comparing the results of \textit{Top n}, \textit{Rand}, and \textit{Bottom}, we demonstrated the importance of sorting images based on the similarity of their old feature. Larger $n$ can lead to somewhat better performance, while as the nearer images should have larger weights according to Section~\ref{sec:collider}, only selecting the nearest image ($n=1$) is also good. Different variants to assign weight values among images, as long as they follow Eq.~\eqref{eq:weight_assignment}, will not affect the final prediction much.

\begin{table}[h]
\centering
\vskip -0.3em
\resizebox{0.9\columnwidth}{8.07mm}{
\begin{tabular}{llllll}
\toprule
\multirow{2}{*}{\textbf{Methods}} & \multicolumn{2}{c}{\textbf{CIFAR-100}} &           & \multicolumn{2}{c}{\textbf{ImageNet-Sub}} \\ \cline{2-3} \cline{5-6}
& \textbf{T = 5}       & \textbf{\quad 10}       & \textbf{} & \textbf{\quad 5}          & \textbf{\quad 10}  \\ \hline
Baseline & 45.57              & 32.72        &     & 58.55               & 45.06               \\
Ours              & \textbf{59.11\footnotesize{\color{red}{+13.54}}  }    & \textbf{55.31\footnotesize{\color{red}{+22.59}}}    & & \textbf{69.22\footnotesize{\color{red}{+10.67}}  }     & \textbf{65.51\footnotesize{\color{red}{+20.45}} }     \\
\bottomrule
\end{tabular}
}
\vskip -0.5em
\caption{Comparisons of average accuracies with no replay data. Baseline is LUCIR.} 
\vskip -0.12em
\label{tab:noreplay}
\vspace{-0.1in}
\end{table}

\begin{table}[t]
\resizebox{\columnwidth}{18mm}{
\begin{tabular}{llccc}
\toprule
& \textbf{Methods}                          & \textbf{R = 5}     & \textbf{10}    & \textbf{20}    \\ \hline
\multirow{4}{*}{\small{Accuracy (\%)}}   & Baseline~\cite{lucir} & 50.76 & 61.68 & 63.57 \\
                                           & \quad +All    & 
                                           59.82\footnotesize{\color{red}{+9.06}} & 
                                           64.41\footnotesize{\color{red}{+2.73}} & 
                                           65.27\footnotesize{\color{red}{+1.70}} \\ \cline{2-5} 
                                           & \quad +DCE      & 58.53\footnotesize{\color{red}{+7.77}} & 64.04\footnotesize{\color{red}{+2.36}} & 65.18\footnotesize{\color{red}{+1.61}} \\
                                           & \quad +MER     & 53.55\footnotesize{\color{red}{+2.79}} & 63.40\footnotesize{\color{red}{+1.72}} & 64.43\footnotesize{\color{red}{+0.86}} \\ \hline
\multirow{4}{*}{\small{Forgetting (\%)}} & Baseline~\cite{lucir} & 28.34 & 17.51 & 14.08 \\
                                           & \quad +All    & 
                                           11.93\footnotesize{\color{red}{-16.41}} & 
                                           6.23\footnotesize{\color{red}{-11.28}}  & 
                                           7.11\footnotesize{\color{red}{-6.97}}  \\ \cline{2-5} 
                                           & \quad +DCE      & 16.82\footnotesize{\color{red}{-11.52}} & 10.16\footnotesize{\color{red}{-7.35}} & 8.41\footnotesize{\color{red}{-5.67}}  \\
                                           & \quad +MER      & 17.45\footnotesize{\color{red}{-10.89}} & 12.15\footnotesize{\color{red}{-5.36}} & 10.01\footnotesize{\color{red}{-4.07}} \\
\bottomrule
\end{tabular}
}
\vspace{-0.1in}
\caption{The individual improvements of the proposed Distillation of Colliding Effect (DCE) and incremental Momentum Effect Removal (MER). Experiments are conducted on CIFAR-100 with 5 incremental steps.}
\label{tab:each_component}
\vspace{0.05in}
\end{table}

\noindent \textbf{Robustness of Incremental Momentum Effect Removal.} In Table~\ref{tb:steps}, we demonstrated the robustness of our incremental Momentum Effect Removal (MER) under the different number of incremental steps. Even in a challenging situation where $N=25$, it still has a significant improvement.

\begin{table}
\resizebox{\columnwidth}{7.5mm}{
\begin{tabular}{c|c|ccc|cc|cc}
\toprule
\textbf{R} & Baseline & Top1 & Top5 & Top10 & Rand & Bottom & Variant1 & Variant2 \\ \hline
\textbf{5} & 50.76 & 55.92 & 58.12 & 58.53 & 54.88 & 41.70 & 58.22 & 58.41 \\
\textbf{10} & 61.68 & 63.51 & 63.66 & 64.04 & 53.80 & 51.41 & 63.54 & 63.97 \\
\textbf{20} & 63.57 & 64.76 & 64.93 & 65.18 & 64.16 & 57.07 & 64.87 & 65.22 \\
\bottomrule
\end{tabular}}
\vspace{-0.1in}
\caption{Comparisons of average accuracies using the Distillation of Colliding Effect (DCE) with different weight assignments. Experiments are conducted on CIFAR-100 with 5 incremental steps. Baseline is LUCIR.}
\label{tab:weight}

\end{table}

\begin{table}
\resizebox{\columnwidth}{13.7mm}{
\begin{tabular}{ccccccc}
\toprule
\multicolumn{1}{l}{} & \textbf{Methods} & \textbf{N=1} & \textbf{2} & \textbf{5} & \textbf{10} & \textbf{25} \\ \hline
 & Baseline~\cite{lucir} & 67.98 & 64.04 & 50.76 & 53.6 & 43.52 \\
\multirow{-2}{*}{Accuracy(\%)} & \ +MER & 68.48 & 65.44 & 53.55 & 57.86 & 45.95 \\ \hline
\rowcolor[HTML]{EFEFEF} 
RI (\%) & \ +MER & 0.74 & 2.19 & 5.50 & 7.95 & 5.58 \\ \hline
 & Baseline~\cite{lucir} & 28.82 & 23.74 & 28.34 & 23.63 & 32.22 \\
\multirow{-2}{*}{Forgetting (\%)} & \ +MER & 25.24 & 16.55 & 17.45 & 11.68 & 20.47 \\ \hline
\rowcolor[HTML]{EFEFEF} 
RI (\%) & \ +MER & 12.42 & 30.29 & 38.43 & 50.57 & 36.47 \\
 \bottomrule
\end{tabular}
}
\vspace{-0.1in}
\caption{Evaluation of average accuracies using incremental Momentum Effect Removal (MER) with varying incremental steps. RI(\%) is the relative improvement in increasing accuracy and lowering forgetting with MER. Experiments are conducted on CIFAR-100  for $R=5$. }
\label{tb:steps}
\vspace{-0.1in}
\end{table}

\noindent \textbf{Different Size of the Initial Task.} As shown in Table~\ref{tab:init_size}, the performance of all the methods decreases due to the degraded representation learned from fewer classes. However,  our  method  still  shows  a  steady  improvement compared to the baseline.

\begin{table}[t]
\centering
\vskip 0.4em
\resizebox{1.0\columnwidth}{12mm}{
\begin{tabular}{lllll}
\toprule
\multirow{2}{*}{\textbf{\begin{tabular}[c]{@{}l@{}}Class size of\\ initial task\end{tabular}}} &
  \multirow{2}{*}{\textbf{Methods}} &
  \multicolumn{3}{c}{\textbf{Average Accuracy(\%)}} \\ \cline{3-5} 
 &
   &
  \multicolumn{1}{l}{\textbf{R = 5}} &
  \multicolumn{1}{l}{\textbf{\quad 10}} &
  \multicolumn{1}{l}{\textbf{\quad 20}} \\ \hline
\multirow{2}{*}{10}  & Baseline~\cite{lucir}            & 50.45               & 54.39               & 58.87               \\
                     & Ours                   & \textbf{52.15\footnotesize{\color{red}{+1.70}}} & \textbf{55.71\footnotesize{\color{red}{+1.32}}} & \textbf{59.17\footnotesize{\color{red}{+0.30}}} \\ \hline
\multirow{2}{*}{20}  & Baseline~\cite{lucir}           & 51.95               & 56.31               & 59.42               \\
                     & Ours                  & \textbf{55.88\footnotesize{\color{red}{+3.93}}} & \textbf{58.38\footnotesize{\color{red}{+2.07}}} & \textbf{59.93\footnotesize{\color{red}{+0.51}}} \\
\bottomrule
\end{tabular}
}
\vskip -0.65em
\caption{Comparisons of average accuracies with the initial task of fewer classes. Experiments are conducted on CIFAR-100 with each incremental step of 10 classes.}
\label{tab:init_size}
\vspace{-0.2in}
\end{table}

\section{Conclusions}
Can we save the old data for anti-forgetting in CIL without actually storing them? This paradoxical question has been positively answered in this paper. We first used causal graphs to consider CIL and its anti-forgetting techniques in a causal view. Our finding is that the data replay, arguably the most reliable technique, benefits from introducing the causal effect of old data in an end-to-end fashion. We proposed to distill the causal effect of the collider: the old feature of a new sample, and show that such distillation is causally equivalent to data replay. We further proposed to remove the incremental momentum effect during testing, to achieve balanced old and new class prediction. Our causal solutions are model-agnostic and helped two strong baselines surpass their previous performance and other methods. In future, we will investigate more causal perspectives in CIL, which may have great potential to achieve the ultimate life-long learning.

{\small
\bibliographystyle{ieee_fullname}
\bibliography{egbib}
}

\end{document}